%% file: template.tex
\documentclass[journal]{vgtc}                     

\onlineid{}

\vgtccategory{Research}

\vgtcpapertype{}

\title{ParetoTracker: Understanding Population Dynamics in Multi-objective Evolutionary Algorithms through Visual Analytics}

\author{%
  Zherui Zhang,
  Fan Yang,
  Ran Cheng, \textit{Senior Member, IEEE},
  and Yuxin Ma, \textit{Senior Member, IEEE}
}

\authorfooter{
  \item All authors are with the Department of Computer Science and Engineering, Southern University of Science and Technology, China. Email: \{zhangzr32021,yangf2020\}@mail.sustech.edu.cn,ranchengcn@gmail.com, mayx@sustech.edu.cn. Z. Zhang and F. Yang contributed equally to this work. Y. Ma is the corresponding author.
}

\abstract{%
Multi-objective evolutionary algorithms (MOEAs) have emerged as powerful tools for solving complex optimization problems characterized by multiple, often conflicting, objectives. While advancements have been made in computational efficiency as well as diversity and convergence of solutions, a critical challenge persists: the internal evolutionary mechanisms are opaque to human users. Drawing upon the successes of explainable AI in explaining complex algorithms and models, we argue that the need to understand the underlying evolutionary operators and population dynamics within MOEAs aligns well with a visual analytics paradigm. This paper introduces ParetoTracker, a visual analytics framework designed to support the comprehension and inspection of population dynamics in the evolutionary processes of MOEAs. Informed by preliminary literature review and expert interviews, the framework establishes a multi-level analysis scheme, which caters to user engagement and exploration ranging from examining overall trends in performance metrics to conducting fine-grained inspections of evolutionary operations. In contrast to conventional practices that require manual plotting of solutions for each generation, ParetoTracker facilitates the examination of temporal trends and dynamics across consecutive generations in an integrated visual interface. The effectiveness of the framework is demonstrated through case studies and expert interviews focused on widely adopted benchmark optimization problems.
}


\teaser{
  \centering
  \includegraphics[width=\linewidth]{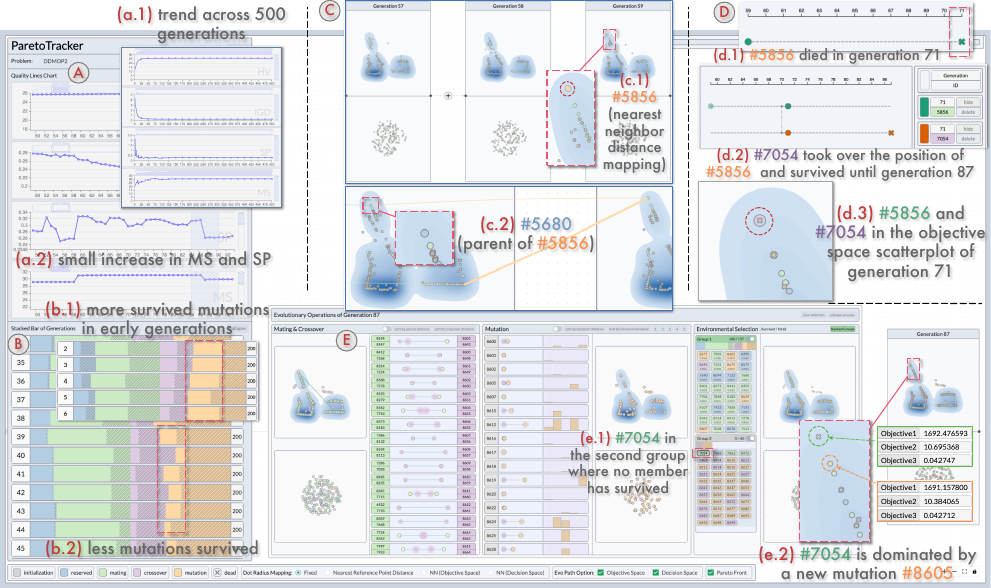}
  \caption{
    The interface of ParetoTracker contains five visualization components: (A) quality line charts, (B) stacked bars for generation statistics, (C) decision and objective space scatterplots for selected generations, (D) the lineage view, and (E) the evolutionary operator view.
  }
  \label{fig:teaser}
}

\graphicspath{{figs/}{figures/}{pictures/}{images/}{./}} 

\usepackage{tabu}                      
\usepackage{booktabs}                  
\usepackage{lipsum}                    
\usepackage{mwe}                       

\usepackage{mathptmx}                  
\usepackage{amsmath}
\usepackage{amsfonts}
\usepackage{enumitem}

\usepackage[table,xcdraw,svgnames]{xcolor}
\definecolor{TODOcolor}{HTML}{0000FF}
\definecolor{AddContentcolor}{HTML}{0000CC}
\definecolor{REVISEcolor}{HTML}{FF0000}
\definecolor{classyellow}{HTML}{FFA600}
\definecolor{classblue}{HTML}{3768B1}
\definecolor{classgreen}{HTML}{167A48}
\definecolor{classpurple}{HTML}{8B79B8}

\newcommand{\MUTATIONCOLOR}[1]{\textcolor{classyellow}{#1}}
\newcommand{\MATINGCOLOR}[1]{\textcolor{classgreen}{#1}}
\newcommand{\RESERVEDCOLOR}[1]{\textcolor{classblue}{#1}}
\newcommand{\CROSSOVERCOLOR}[1]{\textcolor{classpurple}{#1}}

\begin{document}



\input{contents/1_introduction.tex}

\input{contents/2_related_work.tex}
\input{contents/3_overview.tex}
\input{contents/4_framework.tex}
\input{contents/5_evaluation.tex}
\input{contents/6_discussion_conclusion.tex}


\section*{Supplemental Materials}
\label{sec:supplemental_materials}

The supplemental materials include a demo video of our framework and the corresponding subtitle file of the video. An implementation is released at~\url{https://github.com/VIS-SUSTech/ParetoTracker}.

\acknowledgments{%
  This work was supported in part by the National Natural Science Foundation of China (No. 62202217), Guangdong Basic and Applied Basic Research Foundation (No. 2023A1515012889), Guangdong Key Program (No. 2021QN02X794), and Guangdong Natural Science Funds for Distinguished Young Scholar (No. 2024B1515020019).
}

\bibliographystyle{abbrv-doi-hyperref}

\bibliography{template}



\end{document}

%% file: contents/1_introduction.tex
\firstsection{Introduction}

\maketitle

Multi-objective optimization problems frequently emerge in various decision-making scenarios such as engineering design~\cite{pereira2022review}, environmental planning~\cite{arbolino2021multi}, and artificial intelligence~\cite{zhang2021survey}. Unlike single-objective optimization where the goal is to optimize a single criterion, multi-objective optimization considers more than one objectives simultaneously. Since the multiple objectives are often conflicting, improving the performance of one objective might adversely affect others. Consequently, the solution to these problems is not a single optimal point but a set of feasible solutions, termed a \emph{Pareto front}. On this front, each solution is \emph{non-dominated}, indicating that no single solution outperforms others across all objectives consistently.

To address this complexity, the development of multi-objective evolutionary algorithms (MOEAs) has been instrumental. These algorithms leverage evolutionary computation techniques to mediate among the varying objectives~\cite{Tian2021}. MOEAs have emerged as a pivotal approach for solving multi-objective problems over the past decades, especially noted for their capability to manage complex objective functions and to produce a diverse set of solutions in a single algorithm run. The essential mechanism of MOEAs draws inspiration from natural evolution, where only the fittest solutions in terms of optimization objective functions are preserved~\cite{wang2023survey}. In each generation of the evolutionary process, a population of parent individuals\footnote{To facilitate the explanation in this paper, the terms ``individuals'' and ``solutions'' are used interchangeably when discussing evolutionary processes in the context of MOEAs.} are selected to generate offspring individuals through crossover/mutation operations. Thereafter, an environmental selection step filters these individuals, retaining only those with the highest optimization fitness for the next generation.

Despite the efficacy of MOEAs in resolving complex optimization challenges, their operations often remain opaque, which functions as black-boxes to human users~\cite{walter_explainable_2022}. In the realms of evolutionary computing and visualization, efforts have been made to uncover patterns within optimal solution sets~\cite{Filipic2018}. Nonetheless, facilitating user engagement and understanding of the dynamics in evolutionary processes remains largely unexplored, particularly considering the complex evolution strategies employed by various MOEAs~\cite{walter_visualising_2020,walter_visualizing_2022}. Conventional methods of reflecting solution characteristics across generations typically involve numerical quality indicators~\cite{Li2019}, yet a detailed explanation of the evolutionary operations across generations is essential for a comprehensive understanding of how solutions are evolved towards optima. Such insights could empower algorithm developers and practitioners to identify the operational successes or failures of an algorithm as well as gain insights that cannot be gleaned solely through aggregated quality indicators.

The last decade has seen the visualization community successfully apply the visual analytics paradigms to explain complex algorithmic processes, thereby enhancing model transparency~\cite{Hohman2018,yuan2021survey,wang2024survey,LaRosa2023}. We posit that this paradigm is equally applicable to elucidating the evolutionary dynamics within MOEAs. In this paper, we introduce a visual analytics framework, ParetoTracker, designed to facilitate the understanding and inspection of population dynamics in the evolutionary processes of MOEAs. Drawing from an extensive review of relevant literature and collaboration with domain experts in evolutionary computation, we have derived analytical tasks that inform the visualization and interaction design in a hierarchical manner. By employing the ``overview+detail'' scheme, ParetoTracker utilizes a multi-level design that exposes the evolutionary progress across generations from three perspectives: the aggregate measures and statistics of solution sets, cross-generational tracking of individuals and their lineages, and in-depth analysis of evolutionary operations between successive generations. The effectiveness of ParetoTracker is demonstrated with case studies and expert interviews on benchmark problems commonly utilized in the MOEA field.

In summary, our contributions include:
\vspace{-0.7mm}
\begin{itemize}[leftmargin=*]
    \item A visual analytics framework, ParetoTracker, for comprehending and inspecting the dynamics of individuals among generations in the evolutionary processes of MOEAs;
    \item A suite of visual inspection and exploration methods as well as designs to highlight salient evolutionary patterns among generations;
    \item Case studies and expert interviews on established test problems to demonstrate effectiveness of the framework.
\end{itemize}

%% file: contents/2_related_work.tex
\section{Related Work}

Our research addresses the challenges involved in examining and investigating the evolutionary processes in MOEAs. This section provides a review of existing literature on the application of visualization techniques in evolutionary multi-objective optimization algorithms as well as explainable AI.

\subsection{Visualization in Evolutionary Multi-objective Optimization}

Visualization is critical in evolutionary multi-objective optimization due to the complexity of decision and objective spaces, typically exceeding two dimensions. High-dimensional visualization techniques are required to represent decision and objective vectors associated with solutions, such as projection techniques~\cite{pcabook2014,van2008visualizing,mcinnes2018umap}, parallel coordinates plots~\cite{Inselberg2009} (PCPs), and scatterplot matrices~\cite{scatterplotbook1987}. Smedberg and Bandaru~\cite{SMEDBERG20231311} propose an interactive visualization environment using linked scatterplot and PCP as the central representation for solutions. Tu\v{s}ar and Filipi\v{c}'s study~\cite{Tusar2015} on high-dimensional visualization methods highlights their pattern preservation capabilities. They found that projection techniques can support large-scale solution sets and allow simultaneous comparison of multiple sets. However, these techniques may not directly reflect distribution patterns due to potential distortions and two-dimensional result information loss.

The optimization community has also developed specialized techniques for solution set visualization. He and Yen~\cite{He2016} introduce a 2-D radial system for objective vectors. The iSOM method~\cite{Nagar2022,Yadav2023} improves interpretability of self-organizing maps. 3D-RadVis~\cite{Ibrahim2016-3dradvis} shows solution distributions and Pareto front convergence within a single visualization. PalleteViz~\cite{Talukder2020} and PaletteStarViz~\cite{Talukder2020star} highlight geometric properties and constraint boundaries closeness in solution sets.

In additional to the static visualization of algorithm outputs, recent efforts have encompassed the visualization of the evolutionary processes to illustrate the dynamics of the algorithms. De Lorenzo et al.~\cite{de_lorenzo_analysis_2019} and Walter et al.~\cite{walter_visualising_2020,walter_visualizing_2022} employ multi-dimensional scaling (MDS) to visualize the search trajectory within the decision space for single- and multi-objective optimization problems, respectively, which enables the exploration of complex decision space landscape. In the visual analytics community, VisEvol~\cite{chatzimparmpas_visevol_2021} illustrates the internal crossover and mutation mechanisms in evolutionary algorithms in the context of optimizing machine learning parameters. Huang et al.~\cite{Huang2024} propose a comparative analysis framework that utilizes timelines and proximity graphs to investigate the similarities between different algorithm runs.

As such, a large body of research has been dedicated to visualizing solution sets and the dynamics of generations. Nonetheless, a limitation among these methods is their static nature, which offers minimal interactive capabilities. Moreover, these approaches often fall short in offering comprehensive tools for detailed analysis of evolutionary processes at varying levels of granularity. To address this gap, our framework enhances analytical capabilities, enabling evaluations from aggregated metrics down to detailed, fine-grained evolutionary operations.

\subsection{Visual Analytics in Explainable AI}

Over the last decade, a number of surveys~\cite{Hohman2018,yuan2021survey,LaRosa2023,wang2024survey} have summarized advancements in visual analytics applied to machine learning models. Specifically, the visualization community has proposed various techniques and frameworks for ``opening the black box'' of machine learning models~\cite{Endert2017,andrienko2021theoretical}. Work from Muhlbacher et al.~\cite{Muhlbacher2014} summarizes the strategies for presenting information and integrating user controls in black-box models. Kim et al.~\cite{Kim2017} introduce the per-iteration visualization environment (PIVE) to provide insights and enable interactive control in the intermediate iterations. In addition to general methodologies, certain studies have concentrated on providing explanations for white-box models, including tree models~\cite{Lundberg2019,VanDenElzen2011BaobabView,Liu2017b,Zhao2019}, rule-based representations~\cite{Ming2018}, and linear models~\cite{Wang2017,Ma2017svm,Delaforge2023}.

With the evolution of large-scale models such as neural networks, there is a growing necessity to illustrate their training and prediction processes. Usually, the training progress is depicted through statistical charts on a per-iteration basis, which shows key performance metrics like loss or accuracy values~\cite{Chae2017Visualization,Pezzotti2017,Wang2018,Liu2018,Ma2020}. Moreover, there is a body of work dedicated to examining these evolutions through model- or task-specific visualizations, such as DeepTracker~\cite{Liu2018a} for convolutional neural networks, DQNViz~\cite{Wang2018} for deep reinforcement learning, and GAN Lab~\cite{Kahng} for deep generative models.

In summary, the field of explainable AI has witnessed significant research efforts in elucidating the internal mechanisms of complex models. Drawing inspiration from the paradigm of opening black box models characterized by their opaque behaviors, our work focuses on the MOEA domain where the evolutionary processes are difficult to comprehend by human analysts. We tend to uncover the complex evolutionary behaviors hidden in their low-level operations, which provides a more detailed perspective to complement the commonly-used aggregated quality indicators~\cite{Li2019}.

%% file: contents/3_overview.tex
\section{Background}
\label{sec:background}
This section presents a preliminary overview of MOEAs to establish a foundational understanding of the terminology and key components employed in the algorithms.

\begin{figure}[!t]
	\centering	
	\includegraphics[width=1.00\columnwidth]{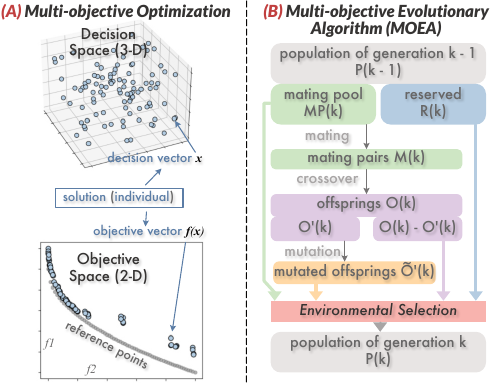}
	\caption{(A) An example of multi-objective optimization problem. (B) The general evolutionary computing pipeline.}
    \label{fig:moea_process}
    \vspace{-4mm}
\end{figure}

\vspace{1.2mm}\noindent\textbf{Multi-objective Optimization.} The multi-objective optimization problem can be formulated as minimizing a series of objective functions:
    \vspace{-1.5mm}
\begin{equation}
    \text{min}\; \mathbf{f}(\mathbf{x}) = (f_1(\mathbf{x}), f_2(\mathbf{x}), \ldots, f_m(\mathbf{x}))
    \vspace{-1.5mm}
\end{equation}

\noindent where $\mathbf{x}$ is an $n$-dimensional \textit{decision vector} $\mathbf{x} = (x_1, x_2, \ldots, x_n)$ in the \textit{decision space} in $\mathbb{R}^n$, and the function evaluation of $\mathbf{x}$ on $m$ objective functions compose the \textit{objective vector} in the \textit{objective space} in $\mathbb{R}^m$, denoted as $\mathbf{f}(\mathbf{x})$, \cref{fig:moea_process} (A). Given that the objectives often show conflicts, e.g., adjusting $\mathbf{x}$ to minimize the value on $f_1(\mathbf{x})$ may increase the outcome on $f_2(\mathbf{x})$, it is unfeasible to find a single solution that simultaneously minimizes all objectives. 

For two solutions $\mathbf{x}_1$ and $\mathbf{x}_2$, $\mathbf{x}_1$ is said to dominate $\mathbf{x}_2$ (termed $\mathbf{x}_1 \succ \mathbf{x}_2$) if and only if $\mathbf{f}(\mathbf{x}_1)$ is no worse than $\mathbf{f}(\mathbf{x}_2)$ on all objectives and strictly better in at least one objective. A solution $\mathbf{x}$ is deemed \textit{Pareto optimal} if there is no other solution in the objective space that dominates $\mathbf{x}$. The set of all Pareto optimal solutions forms the \textit{Pareto Set}, while the corresponding objective vectors of these solutions create the \textit{Pareto Front}. Given such definitions, the primary goal of solving multi-objective optimization problems is to obtain the Pareto.

It should be noted that for benchmarking purpose, some artificial test problems offer \textit{reference sets} that contain numerous objective vectors sampled near the Pareto fronts, \cref{fig:moea_process} (A). This allows comparisons of solutions from certain algorithms with the reference set to assess the  performance of the algorithms.

\vspace{1.2mm}\noindent\textbf{Multi-objective Evolutionary Algorithms.} Among the various strategies for solving multi-objective optimization problems, evolutionary algorithms stand out as one of the most efficient representatives to harvest a solution set in a single run while preserving diversity and convergence~\cite{Li2018}. Starting from a random sample set drawn from the decision space, the MOEAs feature multiple major iterative loops, referred to as \textit{generations}. A loop comprises a fixed number of evolutionary steps, called \textit{operators}, to drive a \textit{population} of candidate \textit{individuals} towards the Pareto front. In the context of MOEAs, each individual corresponds to a solution that encompasses a decision vector.

Shown on the right side of \cref{fig:moea_process} (B), a typical pipeline of an evolutionary algorithm starts from an initialized population $P(1)=\{p_1, p_2, \ldots, p_{\mu}\}$ of size $\mu$, where the $i$-th individual $p_i=(\mathbf{x}_{p_i},\mathbf{f}_{p_i})$ is composed of the decision vector and the corresponding objective vector of the solution. In the iterative evolutions, the $k$-th generation produces a new population of solutions, denoted as $P(k)$, as the evolution outcome. The pipeline inside generation $k$ involves the following operators:

\begin{itemize}[leftmargin=*]
    \item \textit{Mating.} To begin with, the individuals in $P(k-1)$ are firstly divided into two sets: a mating pool $MP(k)$ and a reserved set $R(k)$. Those individuals in $MP(k)$ are then paired into various \textit{mating pairs}: 
    \vspace{-1mm}
    \begin{equation}
        M(k)=\{(p_i, p_j) \mid p_i,p_j \in MP(k), \; i \neq j\}
        \vspace{-1mm}
    \end{equation}
    where $p_i$ and $p_j$ are parents for evolution in the following steps. The number of pairs is denoted as $\lambda$. Note that those \textit{reserved} individuals in $R(k)$ will not join the crossover and mutation operations.

    \item \textit{Crossover.} In this step, each mating pair $(p_i, p_j)$ generates two offsprings by combining characteristics of the parents' decision vectors, namely, $\mathbf{x}_{p_i}$ and $\mathbf{x}_{p_j}$. All the $2\lambda$ new offsprings form an offspring set
    \vspace{-1mm}
    \begin{equation}
        O(k) = \{o_i | 1 \leq i \leq 2\lambda\}
    \vspace{-1mm}
    \end{equation}
    where $o_i = (\mathbf{x}_{o_i}, \mathbf{f}(\mathbf{x}_{o_i}))$ represents the corresponding decision and objective vectors of the offspring.

    \item \textit{Mutation.} A random subset $O'(k)$ of the offspring set $O(k)$, $O'(k) \subseteq O(k)$, is selected, and all individuals in $O'(k)$ will be mutated by applying perturbation operations to the decision vectors, resulting in a mutated offspring set $\tilde{O}'(k)$. In other words, the remaining offsprings $O(k) - O'(k)$ are not mutated.
    
    \item \textit{Environmental Selection.} To determine the optimal individual solutions, a large joint population $Q(k)$ is constructed by joining various population sets from previous steps:
    \vspace{-1mm}
    \begin{equation}
        Q(k) = R(k) \cup MP(k) \cup \tilde{O}'(k) \cup (O(k) - O'(k))
    \vspace{-1mm}
    \end{equation}
    This implies that all the parental individuals (either reserved or selected for mating) from the previous generation, along with their crossover and mutated offsprings, are subject to a competition against one another based on certain selection criteria. Such criteria usually involve measures established on individuals' objective vectors to assess their dominance and diversity~\cite{He2016,Li2018}. Individuals demonstrating a higher contribution to the overall population through optimized objective values or enhancement of the Pareto front coverage will be prioritized. Consequently, the best $\mu$ individuals that prevail in this selection process form the population $P(k)$ as the output of the current generation $k$.
\end{itemize}

The iterative process stops when a predefined termination criterion is satisfied, such as reaching a maximum number of generations. While MOEAs share the abovementioned pipeline for population evolution, they diverge in their specific implementations of selection, crossover, mutation, and environmental selection operators. This divergence in design choices constitutes a primary area of research and development in the evolutionary computation community.

\section{Design Overview}
Leveraging the core principles of MOEAs, we propose a visual analytics framework designed to inspect and explain the evolutionary processes at play within these algorithms. This section outlines the requirements drawn from literature review and collaborative efforts with domain experts in evolutionary computing. Analytical tasks supported by the framework are then derived based on these established requirements.

\subsection{Requirement Analysis}
\label{sec:requirement_analysis}

To accurately identify the requirements from the MOEA domain perspective, we reviewed various surveys~\cite{Li2018,Coello2020,Zitzler2000,Hua2021,Tian2021} and methodological studies~\cite{ALBERTO201433,Filipic2018,de_lorenzo_analysis_2019,walter_visualising_2020,walter_explainable_2022,walter_visualizing_2022} focused on evolutionary process dynamics and visualization in MOEA. This review helped identify potential research gaps. To validate and deepen our understanding, we interviewed two evolutionary computing domain experts, E1 and E2. E1 has over a decade of research experience in MOEAs and their engineering design applications, while E2, a senior PhD student specializing in multi-objective optimization and reinforcement learning, extensively uses evolutionary algorithms. The interviews began with discussions about their MOEAs application practices and experiences with using visualizations to evaluate solution sets. We then discussed the challenges they faced with existing visualization tools and refined the open research questions identified during the literature review. The identified gaps are primarily in two areas:

\vspace{-0.7mm}
\begin{itemize}[leftmargin=*]
    \item \textbf{Limited Native Visualization for Processes.} While widely-used MOEA toolkits~\cite{PlatEMO,Pymoo} incorporate a range of axis- or projection-based visualization techniques for corresponding solution sets from single generations, the process of manually plotting each generation involved in the evolutionary processes proves to be tedious. Moreover, it does not facilitate the examination of dynamics of individual solutions or the trends in populations across iterations due to difficulties in generation-wise comparison.
    \vspace{-1mm}
    \item \textbf{Lack of Inspection Support for Evolutionary Operators.} Another significant limitation in existing toolkits is the absence of visualizations that delve into the level of evolutionary operators. For a detailed analysis of the algorithms, it is essential to explore the mutation and selection operations of specific individuals as well as links to their corresponding parents.
\end{itemize}
\vspace{-0.7mm}

Beyond the two practice gaps, we discussed challenges that hinder the development of tools to bridge these gaps. These challenges fall into three main categories. First, the massive number of individuals involved in the entire evolutionary process can lead to severe visual clutter when using standard visualization techniques. Second, juxtaposing plots from all generations does not assist in identifying temporal patterns or insights due to the typically lengthy iterative process involving hundreds or thousands of generations. The complexity increases when multiple levels of data abstraction, such as the overall statistical level~\cite{Li2019}, generation level, and individual level, are involved simultaneously. Additionally, the varied mechanisms of evolutionary operators among different algorithms make developing unified approaches to assess detailed computation processes difficult. Designing different visualizations for each algorithm can be unrealistic and impractical.

The discussions of gaps and challenges mentioned above yielded essential requirements that our framework should address, ensuring it effectively caters to the needs of researchers and practitioners in the evolutionary computation domain:

\begin{itemize}[leftmargin=*,itemsep=0.04em]
    \item \textbf{R1: Adopt Quality Measures as an Entrance Point.} 
    Using aggregated quality measures as a means to comprehend the performance of MOEAs forms a fundamental basis for algorithm assessment~\cite{Li2019}. E1 suggested that these measures could serve as high-level indicators of potential issues within the populations, thereby prompting a more in-depth analysis.
    \item \textbf{R2: Utilize Multiple Levels of Data Granularity.} Various levels of data granularity, including measure-level, generation-level, solution-level, and operator-level, coexist in the assessment of evolutionary processes. Integrating different levels of granularity into the same environment can facilitate a more comprehensive inspection, encompassing coarse-grained numerical indicators and fine-grained evolutionary operations.
    \item \textbf{R3: Locate and Connect Critical Evolution Behaviors.} Building upon the second requirement, the experts emphasized the significance of identifying representative trends and notable changes across generations. E2 noted the potential for linking observed patterns at the higher level to corresponding detailed evidence at a more granular level, enhancing the depth of analysis.
    \item \textbf{R4: Provide Adequate Abstraction of Operators.} In addition to the challenge of inspecting evolutionary operators, it is crucial to address the need for visualizing diverse implementations of operators. Since MOEAs can vary significantly in their operator design, particularly regarding environmental selection, developing a versatile visualization protocol that accommodates various operator designs is essential for broad applicability.
\end{itemize}

\subsection{Analytical Tasks}
Building on the identified research gaps and requirements, we have distilled the following analytical tasks to guide the framework design:

\vspace{1.2mm}\noindent\textbf{T1: Understand Overall Quality Measures.} Quality measures serve as the primary and most frequently-used indicators for experts and practitioners, which provide a coarse-grained overview of the distribution and performance of solutions in the objective space. This involves:
\vspace{-0.7mm}
\begin{itemize}[leftmargin=*,itemsep=0.08em]
    \item Assessing the solution set quality in a specific generation; (\textbf{R1}, \textbf{R2})
    \vspace{-1.0mm}
    \item Identifying trends within these measures as well as salient changes. (\textbf{R1}, \textbf{R3})
\end{itemize}

\vspace{1.2mm}\noindent\textbf{T2: Explore Dynamics of Individuals among Generations.} At a more granular level, it is essential to uncover the quality of individual solutions and their inheritance relationships across generations:
\vspace{-0.7mm}
\begin{itemize}[leftmargin=*,itemsep=0.08em]
    \item Examining the distribution of solutions across generations and the quality of individual solutions; (\textbf{R1}, \textbf{R2})
    \vspace{-1.0mm}
    \item Tracing the lineage of individuals through multiple generations. (\textbf{R3})
\end{itemize}

\vspace{1.2mm}\noindent\textbf{T3: Inspect Detailed Evolutionary Operations.} At the most detailed level of analysis, there is an interest in understanding the specific actions of evolutionary operations, including:

\vspace{-0.7mm}
\begin{itemize}[leftmargin=*,itemsep=0.08em]
    \item Analyzing the behaviors of evolutionary operators; (\textbf{R3}, \textbf{R4}) 
    \vspace{-1.0mm}
    \item Investigating how an individual from an older generation evolves in the subsequent generation. (\textbf{R3}, \textbf{R4})
\end{itemize}

%% file: contents/4_framework.tex
\section{Visual Analytics Framework}
Based on the identified research challenges, requirements, and analytical tasks, we introduce ParetoTracker, a visual analytics framework designed to illustrate the dynamics of population generations within evolutionary processes of MOEAs with three main components:

\vspace{1.2mm}\noindent\textbf{Performance Overview and Generation Statistics (T1).} As the initial phase of the visual exploration pipeline, this component provides visual representations of algorithm performance alongside statistics of individuals across all generations, \cref{fig:teaser} (A, B). Such design enables analysts to gain a comprehensive understanding of the evolutionary process, identify trends, and discover significant patterns in the measures, thereby guiding further detailed analysis in the following components.

\vspace{1.2mm}\noindent\textbf{Visual Exploration of Individuals among Generations (T2).} After identifying notable sequences of generations in the overviews, detailed information regarding decision and objective vectors is presented in the \textbf{main workspace} of the interface, \cref{fig:teaser} (C). This area facilitates lineage tracing and comparative analysis, supported by an accompanying \textbf{lineage view} shown in \cref{fig:teaser} (D).

\vspace{1.2mm}\noindent\textbf{In-depth Visual Inspection of Operators (T3).} Analysts are allowed to expand the details between two consecutive generations. An \textbf{evolutionary operator view}, \cref{fig:teaser} (E), is provided to depict the intricacies of mating, crossover, mutation, and environmental selection operations on a per-individual basis, thereby offering a nuanced perspective on the evolutionary mechanisms at play. 

\subsection{Performance Overview and Generation Statistics}
\label{sec:performance_overview}

In this component, a suite of quality measures, along with corresponding visualization modules, is utilized to enable a comprehensive analysis of the generations (\textbf{T1}, \textbf{R1}). Additionally, statistics regarding the type of evolutionary operations are provided to offer insights into the evolutionary processes occurring in each generation.

\vspace{1.2mm}\noindent\textbf{Generation-level Quality Measures.} Drawing from established literature on the quality assessment of solution sets in multi-objective optimization~\cite{Li2019}, the following four quality measures have been selected to evaluate generations. These measures fall into two categories: aggregated measures, which assess multiple aspects of quality simultaneously, and specialized measures, which focus on a specific aspect.

\begin{itemize}[leftmargin=*]
    \item \textit{Inverted Generational Distance (IGD) and Hypervolume (HV)}: IGD and HV are among the most frequently-used aggregated measures in the MOEA literature by simultaneously evaluating both convergence and diversity. IGD calculates the average distance from the solutions in the population to the points in the reference set in the objective space, whereas HV quantifies the volume of the region dominated by the population relative to a predefined point in the objective space. A lower IGD or a higher HV value signifies a population with superior convergence and diversity.
    \vspace{-1.0mm}
    \item \textit{Spacing (SP) and Maximum Spread (MS)}: In contrast to the aggregated quality indicators, SP and MS focus on specific aspects of the quality of solution sets. SP computes the uniformity of the solutions by measuring the variance in distances between them, with a larger SP value indicating a less uniform distribution of solutions. MS evaluates the extent of the solutions across each objective, with a larger MS value denoting a broader spread of solutions.
\end{itemize}

As depicted in \cref{fig:teaser} (A), the selected quality measures are visualized using four separate quality line charts in the quality line chart view, each corresponding to one measure. The horizontal axis, aligned across all line charts, denotes the sequence of generations, while the vertical axis represents the corresponding measure values. A semantic zoom feature is implemented on the horizontal axis for detailed examination of specific generation ranges, and the zoom ratios are synchronized to support alignment and comparison across all line charts.

\vspace{1.2mm}\noindent\textbf{Statistics of Populations.} Understanding the origins of individuals concerning evolutionary operations is vital for analyzing population dynamics. Specifically, analysts need to discern whether an individual in a given generation is a directly inherited parent from the previous generation or a mutated offspring evolved from a parental pair. To address this, we employ a stacked bar design visualization to depict individual proportions based on their origins. As shown in \cref{fig:teaser} (B), the ordinal indices of generations are placed on the stacked bars' left side. Inside the bar for the $k$-th generation, a series of segments represent the proportions of the following four types of origins:

\vspace{-1.5mm}
\begin{enumerate}[label=\arabic*),leftmargin=1.7em,noitemsep]
    \item Individuals reserved from the previous generation, $R(k)$;
    \item Individuals in the mating pool, $MP(k)$;
    \item Crossover offsprings based on their parental pairs, $O(k) - O'(k)$;
    \item Mutated offspring, $\tilde{O}'(k)$.
    \vspace{-1.5mm}
\end{enumerate}

A categorical color scheme is employed to facilitate distinction among the four origins, where light blue is for \RESERVEDCOLOR{reserved individuals}, green for \MATINGCOLOR{individuals from the mating pool}, purple for \CROSSOVERCOLOR{crossover offspring}, and yellow for \MUTATIONCOLOR{mutated offspring}. It is worth-noting that this color mapping is consistently applied in the entire framework. Additionally, bar segments featuring striped patterns indicate the proportions of individuals that did not survive the environmental selection process.

\subsection{Population Analysis of Generation Sequences}
\label{sec:population_analysis}

Delving into the specifics of solutions, including decision and objective vectors of individuals across all generations, facilitates a granular understanding of evolutionary dynamics beyond mere aggregate measures and statistics (\textbf{T2}, \textbf{R3}). This component leverages an abstraction of individuals and their lineages, displayed in the main workspace of the interface, alongside a lineage view, \cref{fig:teaser} (C, D).

\vspace{1.2mm}\noindent\textbf{Decision and Objective Space Analysis.} When notable patterns like salient increases or decreases in quality measure values are discovered, analysts can select a range of generations by brushing along the horizontal axis or clicking on any data point in quality line charts. The detailed information for the selected generations is then depicted in a juxtaposed manner in the main workspace, \cref{fig:teaser} (C). Following \textbf{R4} for providing appropriate data abstraction at each analytical level, solutions are abstracted as two sets of vectors, i.e., decision and objective vectors, and are visualized in two vertically-stacked scatterplots, with the top one for the objective space and the bottom one the decision space.

\vspace{1.1mm}\noindent\textit{Layout}: It is common for the dimensions of decision and objective spaces to exceed two, sometimes reaching beyond ten. To address this, projection techniques are employed, utilizing PCA for the objective space and t-SNE~\cite{van2008visualizing} for the decision space. To ensure comparability, we address the alignment of projection results within the same space:

\vspace{-1.0mm}
\begin{itemize}[leftmargin=*,itemsep=0.08em]
    \item For the objective space, the PCA projection matrix is fit using the reference set and subsequently applied across all scatterplots, enabling clear observation of solution distributions relative to the ground-truth reference points. 
    \vspace{-1.0mm}
    \item For the decision space, t-SNE parameters are trained based on the union of all decision vectors present in the selected generations, with these parameters consistently applied across all selected generations.
\end{itemize}
\vspace{-1.0mm}

It should be noted that the choice of the projection techniques is based on the typical configuration in multi-objective optimization problems, where the count of objective functions generally does not exceed 10, often hovering around 4. Conversely, the dimensionality of decision space can span from tens to thousands~\cite{Tian2021}. Given this disparity, linear projection techniques may fall short in adequately unveiling the complex distributions and patterns in the decision space.

\vspace{1.1mm}\noindent\textit{Visual Encoding}: The dot colors are mapped to the origins of evolutionary operations based on the abovementioned categorical color scheme in \cref{sec:performance_overview}. To highlight the quality of individuals, analysts can assign dot sizes to the values of two individual-level quality measures: 1) the distance to the nearest reference point~\cite{van1998evolutionary}, and 2) the distance to the nearest solution in the objective space. The former size mapping serves as a visual cue for the convergence of solutions towards the Pareto front, i.e., the accuracy of the solutions, and the latter one illustrates the uniformity of distributions, with smaller dot sizes implying shorter distances. If clusters of small-sized dots are observed, it strongly suggests that the corresponding regions may contain a considerable number of duplicated solutions. On the other hand, large and scattered dots indicate regions that are relatively sparse. Besides the objective space, the decision space scatterplots incorporate an additional size mapping mode by utilizing the distance to the nearest solution within the decision space, which further reveals the spatial relationships among solutions in the decision space. To illustrate the outcomes of the environmental selection process for each generation, individuals that do not survive are distinctly marked as crosses instead of dots in the scatterplots and will not appear in successive generations.

For test problems with reference sets, a straightforward visualization approach is to plot the corresponding objective vectors of reference points in the objective space scatterplot. However, this can lead to significant visual clutter due to the high number of sampled reference points typically involved~\cite{He2016,Huang2024}. To address this, we adopt a density map to illustrate the distribution of the reference points. The density map is positioned beneath all marks representing individuals to provide an unobstructed view, effectively enabling comparisons between the current population spread and the desired Pareto front.

\vspace{1.1mm}\noindent\textit{Interaction}: When hovering the mouse pointer over a dot, a tooltip can be activated to display critical attributes about the solution. The information includes the generation order, individual ID, the origin of evolutionary operator, the distance to the nearest reference point distance, the values on all objectives, and the survival status after the environmental selection process in the corresponding generation.

\vspace{1.2mm}\noindent\textbf{Lineage Analysis.} Beyond simply placing scatterplots of all selected generations side by side, our framework also facilitates a detailed depiction of the parental relationships between generations. Selecting any dot in a scatterplot triggers the display of connecting curves that link the individual either to its parents (or back to itself if it is a direct inheritance from the preceding generation), \cref{fig:teaser} (C). These parental links extend all the way to the initial generation in the chosen range. The color of the curves corresponds to the offspring's origin of evolutionary operation, while the curve thickness illustrates the objective space distance between the offspring and the linked parent. Enable the lineage connections for multiple individuals simultaneously is also supported.

\begin{figure}[!t]
	\centering	
	\includegraphics[width=1.00\columnwidth]{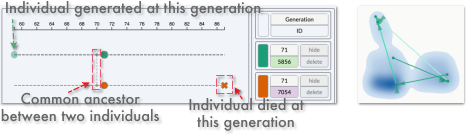}
	\caption{Lineage view. (Left Panel) A timeline of the selected individuals. (Right Panel) A scatterplot depicting the orders of ancestors with arrows.}
    \label{fig:lineage}
    \vspace{-4mm}
\end{figure}

Analysts might be curious about a group of individuals to determine their long-term evolutions as well as whether they evolved from a common set of ancestors. However, visualizing lineage trees in scatterplots across a considerable amount of generations may lead to visual clutter, especially when comparing multiple trees for several individuals. To address this, a dedicated \textit{lineage view} is designed to clarify the illustration of multiple lineages, \cref{fig:teaser} (D) and \cref{fig:lineage}, with two panels:

\vspace{1.1mm}\noindent\textit{Left Panel}: The horizontal axis represents the sequential order of generations, with each lineage-enabled individual represented as a timeline in a row. The timeline highlights the originating generation of the individual and the point at which it no longer passes the environmental selection process. For illustrating common ancestors among individuals, cross-row curved links are used for showing the generation where a closest common ancestor is identified. Note that the range of generations covered by the horizontal axis only spans the necessary range of the selected individuals instead of depicting all generations in the evolutionary process. The axis starts from the earlist generation where a common ancestor appears or an individual is born, and it ends with the final survived generation among all selected individuals.

\vspace{1.1mm}\noindent\textit{Right Panel}: In addition to the temporal comparisons, this panel visualizes the distribution of the selected individual and its lineage across the generations on the two scatterplots for decision and objective vectors, respectively. Arrows pointing from parents to offsprings are rendered in gradient shades where lighter tones indicate earlier generations. This assist in examining the decision and objective vector distributions of all related individuals on the lineage tree collectively.

\subsection{In-depth Inspection of Evolutionary Operations}
At the most detailed level of analysis, this component offers insights into the specific operations executed on the population within one generation to create the  subsequent generation. When a range of generations is selected in the main workspace, clicking on the ``plus'' sign between the scatterplots of two consecutive generations unfolds an \textit{evolutionary operator view} which shows the intricate steps of each evolutionary operation. Drawing from the foundational principles of evolutionary computing outlined in~\cref{sec:background}, we specify these operations within the multi-objective optimization context, aiming to provide data abstraction that aligns with analysts' requirements as suggested in \textbf{R4}.

\vspace{1.2mm}\noindent\textbf{Data Abstraction for Operators in MOEAs.} Through discussions with domain experts and an examination of prevalent MOEA tools like PlatEMO~\cite{PlatEMO} and Pymoo~\cite{Pymoo}, we have identified detailed abstractions for mating, crossover, mutation, and environmental selection operations, encompassing a broad spectrum of MOEAs.

\vspace{1.1mm}\noindent\textit{Mating and Crossover}: The common mating protocol used in MOEAs is the same as how $M(k)$ is constructed in~\cref{sec:background}. For crossover, we focus on the Simulated Binary Crossover (SBX) operator~\cite{deb2007self}, tailored for real-valued recombination operations. Given a pair of individuals $(p_i, p_j)$ with their decision vectors $\mathbf{x}_{p_i}$ and $\mathbf{x}_{p_j}$, SBX yields two offsprings $o_i$ and $o_j$ with decision vectors calculated as follows:
\vspace{-2.0mm}
\begin{gather}
    \mathbf{x}_{o_i} =  0.5 \left[ (1 + \beta) \mathbf{x}_{p_i} + (1 - \beta) \mathbf{x}_{p_j} \right]\\
    \mathbf{x}_{o_j} =  0.5 \left[ (1 - \beta) \mathbf{x}_{p_i} + (1 + \beta) \mathbf{x}_{p_j} \right]
\end{gather}

\vspace{-1.8mm}
where $\beta$ is a spread factor derived from a specific probability distribution. In a more intuitive explanation, this approach linearly combines the parents' decision vectors with symmetric random perturbations, and all four vectors are along the same line in decision space. Notably, some implementations~\cite{Pymoo} introduce an additional post-crossover random perturbation to enhance the genetic diversity of the offsprings.

\vspace{1.1mm}\noindent\textit{Mutation}: Alongside SBX, polynomial mutation is frequently employed. For each selected offspring in $O'(k)$ undergoing mutation, a random perturbation vector $\delta \in \mathbb{R}^n$ is generated in a defined range to avoid excessive and unrealistic modifications.

\vspace{1.1mm}\noindent\textit{Environmental Selection}: At a high level, this process involves labeling candidate individuals as ``survived'' or ``died''. Yet, this abstraction may not fully capture the nuanced selection mechanisms inherent in MOEAs. After exploring various search and selection strategies of algorithms~\cite{wang2023survey}, we suggest a more intricate two-level abstraction: grouping and fitness scoring.
\vspace{-0.2mm}
\begin{itemize}[leftmargin=*,itemsep=0.08em]
    \item Candidates are initially sorted into groups based on specific criteria. Strategies like non-dominated sorting~\cite{deb2002fast} prioritize groups so that individuals in lower-priority groups are dominated by at least one individual in higher tiers. Conversely, methods like uniform space partitioning~\cite{cheng2016reference} treat all groups as equal in priority.

    \vspace{-1.0mm}

    \item Within these groups, individuals receive a fitness score that dictates their competitive standing for survival. As mentioned in \cref{sec:background} on environmental selection operators, such a fitness score reflects an individual's contribution to the population in terms of its proximity to the Pareto front and its coverage on the Pareto front. For instance, NSGA-II~\cite{deb2002fast} uses crowding distances to eliminate duplicated solutions in densely populated regions while favoring solutions in less crowded areas, thereby enhancing population diversity. MOEA/D~\cite{zhang2007moead} ranks solutions using a weighted-sum method over objective values in their corresponding objective vectors. However, the exact method for computing fitness scores can differ across various MOEAs, leading to our design choice of using numerical scores rather than illustrating the detailed computation process. Note that in the prioritized grouping settings, the primary determinant is group rank, followed by individual fitness scores. For non-prioritized groupings, quotas for survival are assigned in each group based on fitness scores.
\end{itemize}

\vspace{-0.05mm}\noindent\textbf{Visualization and Interaction.} The evolutionary operator view is divided into three distinct panels, each corresponding to one of the evolutionary operations previously discussed. The basic layout for these panels includes a list of individuals alongside a pair of scatterplots for the decision and objective spaces. However, each panel features a tailored design to effectively represent the details of different operators.

\begin{figure}[!t]
	\centering	
	\includegraphics[width=1.00\columnwidth]{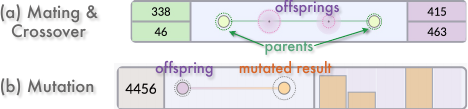}
	\caption{Examples of the glyphs for (a) the mating and crossover panel and (b) the mutation panel.}
    \label{fig:glyphs}
    \vspace{-4mm}
\end{figure}

\vspace{1.1mm}\noindent\textit{Mating and Crossover Panel}: This panel displays each pair of parents from the previous generation and their offsprings in a list format where a glyph design, \cref{fig:glyphs}, is employed to illustrate the relationships between parents and offsprings. In the glyph, two green dots represent the parents, and two tiny purple dots represent the offspring. The connecting line's length between the two parent dots signifies the Euclidean distance between the parent individuals in the decision space. A Gaussian-blurred halo around the purple offspring dots indicates the degree of post-hoc random perturbations applied to the offspring before mutation, with a larger halo showing a greater perturbation. Dashed concentric circles around the dots denote the quality of the individuals, indicating whether the crossover operation has produced offsprings superior to the parents. The radii of these circles are proportional to the distance to the nearest reference point, which is the same individual-level quality measure utilized in~\cref{sec:population_analysis}. The scatterplots for decision and objective spaces display the spatial distributions of the parents, the reserved individuals from the previous generation, and the offspring within these two spaces. The rows in the list can be sorted by parent-parent or parent-offspring distances.

\begin{figure*}[!t]
	\centering	
	\includegraphics[width=1.0\textwidth]{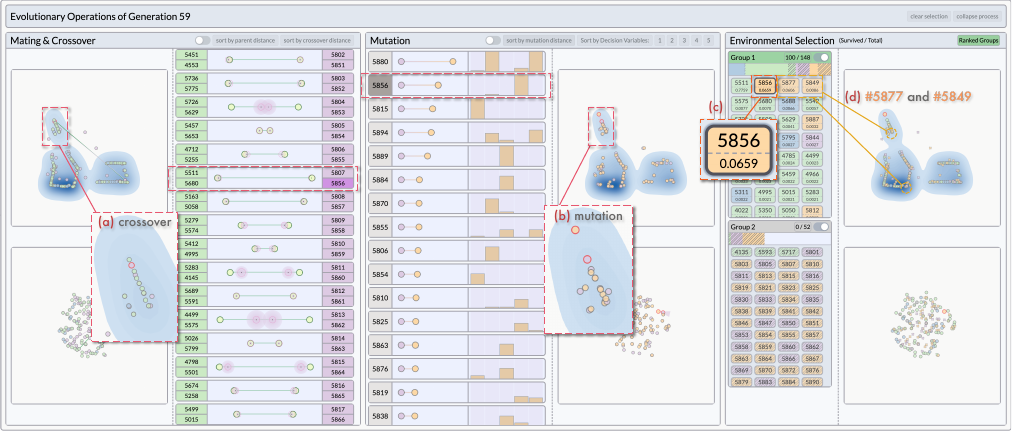}
	\caption{The evolutionary operator view for the 59th generation.}
    \label{fig:case1_operator}
    \vspace{-5mm}
\end{figure*}

\vspace{1.1mm}\noindent\textit{Mutation Panel}: A similar glyph design is used to list all mutated offsprings, \cref{fig:glyphs} (b), where purple and yellow dots represent the crossover offsprings and its mutated counterparts, respectively. The linkage length between dots illustrates the Euclidean distance the offspring has mutated in the decision space. The same dashed concentric circles around the dots convey the individual's quality, specifically its convergence to the Pareto front. To illustrate the direction of mutations in the decision space, a bar chart is utilized for each mutation to display the values of all dimensions in the mutation direction vector, represented as $\mathbf{x}_{\text{mutation}} - \mathbf{x}_{\text{offspring}}$ in $\mathbb{R}^n$. The values along the same dimension are subjected to min-max normalization to [0, 1], facilitating a straightforward comparison across different mutations and dimensions. List sorting is supported based on different criteria, including distances between offsprings and corresponding mutated results as well as dimensions of the mutation direction vectors.

\vspace{1.1mm}\noindent\textit{Environmental Selection Panel}: Individuals entering the environmental selection process are initially sorted into groups, with each group displayed in the list according to their priority. Individual IDs, evolutionary operation origins (indicated by the background color), and fitness scores are depicted in small rectangles within each group. The individuals in groups are further organized by fitness scores, highlighting the top-performing individual. The origin proportions of evolutionary operations within each group are represented as stacked bars, consistent with the design for generation statistics in~\cref{sec:performance_overview}. The total length of the stack bars across all groups is normalized based on the number of individuals in the groups, with the total and survived individual counts displayed at the group's header.

\vspace{1.2mm}\noindent\textbf{Interactions.} The evolutionary operator view facilitates rich interactions. Clicking on any list row or scatterplot dot highlights corresponding individuals across the entire view; in addition, linking lines between dots are displayed simultaneously to visualize relationships, including parental links in the mating and crossover panel and the connection between offspring and their mutations. Toggle button in the group headers enables simultaneous highlighting of all individuals in a group. The tooltip feature for dots of individuals, as discussed in \cref{sec:population_analysis}, is also applied across the scatterplots in the panels.

%% file: contents/5_evaluation.tex
\section{Evaluation}

In this section, we discuss the results from case studies and expert interviews to demonstrate the effectiveness of ParetoTracker on benchmarking and real-world multi-objective optimization problems. ParetoTracker is implemented with a browser-server architecture by employing Python Flask for the server-side logic as well as Vue3 and D3.js for the client side. Pymoo~\cite{Pymoo} is used for running MOEAs on test problems.

\subsection{Case Study 1: SMS-EMOA on DDMOP2 Problem}
In the first case study, we employ the DDMOP2 test problem from the DDMOP test suite~\cite{He2020}, where the problems are devised from scenarios encountered in real-world applications. The DDMOP2 problem comprises five decision variables which represent the reinforcing components of automobile frontal structures while aiming to minimize three objectives. The SMS-EMOA algorithm~\cite{beume2007sms} is selected for this task due to its prominence among indicator-based MOEAs. SMS-EMOA prioritizes individuals based on their contribution to the Hypervolume (HV) measure, which is indicative of an individual's importance in the environmental selection process, and such approach is known for ensuring good coverage and uniformity along the Pareto front. The population size is set to 100, and the algorithm run lasts 500 generations.

\vspace{1.2mm}\noindent\textbf{Quality Measures and Generation Statistics (T1).} After the data is loaded, an initial review of quality line charts and generation stacked bars provides an overview of the evolutionary process, \cref{fig:teaser} (a.1). The two measures, HV and IGD, show relatively monotonic trends towards their desired optimality, which is characterized by increasing HV and decreasing IGD values. Meanwhile, SP and MS display fluctuations in the early evolutionary stages (up to the 40th generation) before stabilizing. Typically, changes in SP and MS are strong indicators that the coverage and spread of the population on the Pareto front are still expanding, a phenomenon primarily triggered by mutated individuals. Following the 40th generation, these measures begin to stabilize, suggesting that little to no effective mutations are occurring, and the solution distribution is starting to reach a steady state. Such stabilized status can be confirmed by the stacked bars, which show a decreasing number of survived mutated individuals after the 40th generation, \cref{fig:teaser} (b.1, b.2). Notably, in the quality line charts, a small increase in MS is observed between the 58th and 87th generations, accompanied by a similar value rise in SP, indicating noteworthy variations in solution distributions that requires closer examination, \cref{fig:teaser} (a.2).

\vspace{1.2mm}\noindent\textbf{Generation-level Analysis (T2).} Triggered by the notable increase in MS and SP, a detailed examination of solution distributions is conducted by activating the scatterplot series around the 58th generation. After clicking on the point representing the 58th generation in the MS line chart, scatterplots for three consecutive generations both prior to and following the 58th generation can be displayed in the main workspace, \cref{fig:teaser} (C). In the results, an important observation between the 58th and 59th generations is the emergence of a mutated individual (ID \MUTATIONCOLOR{\#5856}) in the 59th generation, which broadens the solution set coverage in the top left corner of the objective space scatterplot relative to the blue density map of reference points, \cref{fig:teaser} (c.1). This mutation contributes to the MS increase by expanding the range of the solution set with respect to values covered in the objective space. Additionally, by employing the dot size mapping based on nearest neighbor distances in the objective space, it can be observed that the dot size of {\#5856} is significantly larger than most other individuals, suggesting its role in contributing to the increase of SP value observed in the line chart.

Upon further investigation by clicking on the dot of {\#5856}, lineage connections displayed in the preceding scatterplots reveal that {\#5856} emerged from a mutation involving a pair of parents, one of which, \RESERVEDCOLOR{\#5680} from the 58th generation, is extremely close to the offspring {\#5856} in the objective space (\cref{fig:teaser} (c.2)). The timeline in the lineage view indicates that {\#5856} does not persist beyond the 71st generation, \cref{fig:teaser} (d.1). Nonetheless, the MS and SP measures remain elevated until the 87th generation. This observation implies that other solutions in the vicinity of {\#5856} may have taken over its role in the solution space. This hypothesis is validated by examining scatterplots around the 71st generation, where another individual, \CROSSOVERCOLOR{\#7054}, appears in nearly the identical location as {\#5856}, \cref{fig:teaser} (d.2, d.3).

\vspace{1.2mm}\noindent\textbf{Evolutionary Details between Genrations (T3).} To conduct a fine-grained analysis, we delve into the specifics of how {\#5856} emerged in the 59th generation and the circumstances leading to the demise of {\#7054} in the 87th generation. By activating the evolutionary operator view for the 59th generation and selecting {\#5856} in the objective scatterplot in the environmental selection panel, all elements related to {\#5856} are highlighted for analysis, \cref{fig:case1_operator}.

In the mating and crossover panel, the offspring from the crossover operation does not initially occupy the position of {\#5856}, \cref{fig:case1_operator} (a). Instead, it is the subsequent mutation operation that propels the offspring towards the top left part of the cluster, thus shaping the distinctive characteristics of {\#5856}, \cref{fig:case1_operator} (b). In the grouped list of the environmental selection panel, {\#5856} ranks second in fitness score in the first group, \cref{fig:case1_operator} (c), reflecting its significant contribution to the HV indicator in SMS-EMOA and, by extension, to the overall solution set quality. This observation extends to other mutated individuals like \MUTATIONCOLOR{\#5877} and \MUTATIONCOLOR{\#5849}, \cref{fig:case1_operator} (d), who exhibit a similar expansion in solution set coverage at critical cornering locations. Such patterns suggest a possible algorithmic preference for individuals enhancing Pareto front coverage.

Turning to evolutionary operator view of the 87th generation, \cref{fig:teaser} (e.1), the environmental selection panel reveals that {\#7054} falls in the second group where all members did not survive. This situation results from the non-dominated sorting mechanism in SMS-EMOA, where {\#7054} is outperformed and thus dominated by at least one individual from the prioritized first group. A closer inspection of objective vector values in \cref{fig:teaser} (e.2) indicates that {\#7054} is dominated on all objectives by a mutated individual, \MUTATIONCOLOR{\#8605}, leading to its elimination from the evolutionary process. This analysis reveals the dynamic interplay of evolutionary operators and selection pressures that shape the development and eventual pruning of solutions in MOEAs.

\begin{figure}[!t]
	\centering	
	\includegraphics[width=1.0\columnwidth]{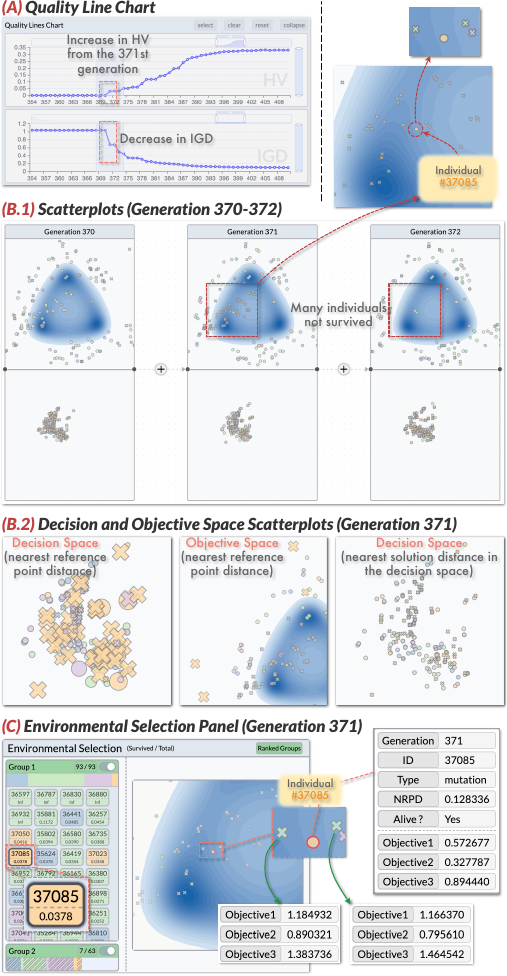}
	\caption{The result of NSGA-II on DTLZ3. (A) The HV measure starts to increase from the 371st generation. (B.1) Individual \#37085 in the 371st generation has dominated a considerable amount of nearby individuals. (B.2) The distributions of decision and objective vectors exhibit a large incoherence. (C) The details of how \#37085 dominates other individuals are illustrated.}
    \label{fig:case2}
    \vspace{-6mm}
\end{figure}

\subsection{Case Study 2: NSGA-II on DTLZ3 Problem}

The second case study explores the DTLZ3 problem from the widely-used DTLZ benchmark suite for multi-objective optimization~\cite{Deb2002DTLZ}, which features 10 decision variables and 3 objectives. The NSGA-II~\cite{deb2002fast} algorithm is selected for this analysis. We adopt the same population and generation settings as in the first case study.

\vspace{1.2mm}\noindent\textbf{Quality Measures and Generation Statistics (T1).} Observations from the quality line chart and generation stacked bars, \cref{fig:case2} (A), highlight a notable uptick in the HV measure at the 371st generation, alongside an accompanying drop in IGD pointing towards an optimal trend. This pattern strongly suggests that the population may have begun converging on the Pareto front after an extensive exploration phase where the measures do not vary too much; prior to the population converging on the Pareto front, it is likely that it remains considerably distant from the target Pareto front.

\vspace{1.2mm}\noindent\textbf{Generation-level Analysis (T2).} The notable shifts observed in the HV and IGD metrics necessitate a closer examination of the scatterplot series around the 371st generation, \cref{fig:case2} (B.1). Prior to the 371st generation, a cluster of individuals is situated near the bottom-left part of the reference point density map. A critical transition occurs in the 371st generation, where the majority of these individuals do not survive the environmental selection process, leaving behind only one mutated survivor, \MUTATIONCOLOR{\#37085}. From the 372nd generation onwards, the emergence of new crossover and mutated individuals is noted in this region.

When switching the dot size mapping to the nearest reference point distances in the enabled scatterplots, \cref{fig:case2} (B.2), an abundance of yellow crosses in the decision space scatterplots have appeared, with their counterparts rarely visible in the objective space scatterplots. Panning the viewport of the objective space scatterplots uncovers these crosses positioned far from the reference density map area. This observed phenomenon contrasts sharply with the distribution patterns in the t-SNE projection results in the decision space scatterplots, where individuals appear to be relatively uniformly distributed. Such uniformity can further be confirmed when applying dot size mapping to the nearest solution distances in the decision space, where the dot sizes have very few variations. The discrepancy between the uniform distribution in the decision space and the varied distribution in the objective space underscores the high sensitivity of the objective functions to the input decision vectors. In other words, small changes in the decision space can lead to significant differences in the objective space, which suggests high complexity of the decision space landscape~\cite{Munoz2015}.

\vspace{1.2mm}\noindent\textbf{Evolutionary Details between Genrations (T3).} Further examination is directed towards {\#37085} in the 371st generation to understand its survival, \cref{fig:case2} (C). Highlighting {\#37085} in the environmental selection panel for this generation reveals its placement in the first group based on the non-dominated sorting mechanism in NSGA-II. Examination of the nearest reference point distances via tooltips shows that {\#37085} exhibits significantly lower values, with its objective vector values also lower across all objectives compared to surrounding non-surviving individuals. These insights confirm {\#37085}'s superiority over its nearby counterparts, ensuring its preservation within the population.

\subsection{Expert Interview}
The visual analytics framework was further evaluated through interviews with four domain experts, including those two from the preliminary interviews (E1 and E2) and two additional experts, denoted as E3 and E4, respectively. E3, a senior researcher, has extensive experience in developing optimization algorithms for real-world applications. E4, a graduate student, concentrates on multi-objective optimization research. The interview process began with an introduction to the framework and a demonstration of its analytical features. Then, we encouraged the experts to freely explore the system using pre-loaded data corresponding to the two case studies. They were also guided to go through the analysis processes that led to the results presented in the case studies. Feedback was collected on findings in case studies, functionality for data analysis, and visualization design of the framework. Interviews lasted between 30 minutes to 1.5 hours.

\vspace{1.2mm}\noindent\textbf{Experts' Comments on Case Studies.} The experts provided several pieces of feedback regarding the observed evolutionary processes in the case studies. For the DDMOP2 problem, where a mutated individual \#5856 initiated an exploration in the objective space to enhance the population's coverage but ultimately failed, experts posited two possible explanations. First, the algorithm's characteristics may align perfectly with the nature of the test problem, leading to the population converging to the Pareto front around the 40th generation and stabilizing. In this scenario, mutations like \#5856 are unnecessary. However, the experts emphasized that this conclusion should be assessed by comparing runs from different algorithms to further evaluate the convergence. Second, the algorithm may discourage mutations that increase diversity and consistently generate individuals with higher accuracy, even if they are duplicates in certain local areas. The experts suggested that improvements could be made by adjusting environmental selection strategies in the next step to see results under different mutation tolerances.

Conversely, mutation \#37085 in the DTLZ3 problem illustrates a successful exploration towards the Pareto front, which results in the ultimate success of convergence despite being continually poor according to the quality line charts (starting to converge only after the 371st generation). The experts noted that it could be an interesting topic to further investigate the underlying factors that generate successful mutations, as this could inform future algorithm design.

\vspace{1.2mm}\noindent\textbf{Overall Analysis Pipeline.} The experts expressed enthusiasm for the incorporation of interactive visualization techniques to examine evolutionary processes in MOEAs. They commended the capacity of the framework to represent generations and individuals across various aspects, including quality measures, generation statistics, and detailed process views. E1 highlighted the advantage of ParetoTracker's multi-faceted design over traditional systems that rely on static visualization, noting its facilitation of interactive exploration and pattern discovery across generations. E4 recognized the potential of ParetoTracker for integrating into daily research workflows, praising its adaptability for comprehensive analysis of different algorithms. ``The proposed data abstraction method accommodates a majority of MOEA algorithms. By adapting the logging mechanisms of evolutionary computing processes, ParetoTracker could be seamlessly connected to computational tools once the output meets the data protocol,'' remarked E4. 

\vspace{1.2mm}\noindent\textbf{Visualization and Interaction.} The interactive visualization design received positive feedback for its efficacy in depicting various measures, solution distributions, and the environmental selection process across generations. E2 mentioned that patterns identified at higher analytical levels could serve as cues for subsequent detailed investigations, thereby enriching the insights obtained. E3 emphasized the value of a detailed view of evolutionary operations, stating, ``Sometimes the algorithm gets stuck in suboptimal conditions as indicated by quality measures. The individual-level visualization, which illustrates the creation and disappearance of individuals, can help identify pivotal evolutionary events that either enhance or hinder optimization efforts.''

\vspace{1.2mm}\noindent\textbf{Suggestions on Improving the Current Framework.} The experts offered suggestions for potential future improvements to the framework. E4 suggested extending the data abstraction protocol for evolutionary operations to include mechanisms for introducing new, randomly selected individuals into the population, thereby increasing diversity. E3 recommended incorporating contextual information or visualizations related to the specific application domains of test problems, which could enhance the applicability of the framework in particular scenarios.

%% file: contents/6_discussion_conclusion.tex
\section{Discussion and Conclusion}

We present ParetoTracker, a visual analytics framework illustrating the evolutionary dynamics in multi-objective evolutionary algorithms. Its multi-level design allows analysts to explore evolutionary processes from different angles, including quality measures, generation statistics, solution distributions across generations, and evolutionary operator actions. This holistic approach enables analysts to go beyond conventional single solution set analysis, promoting a deeper exploration into the evolutionary operations of each generation.

\vspace{1.2mm}\noindent\textbf{Comparison with Existing Tools.} Our framework is compared with existing MOEA frameworks and toolkits~\cite{Pymoo,PlatEMO}, focusing on evolutionary dynamics analysis support. These major computation toolkits offer static visualization for quality measures and solution sets through non-interactive charts and scatterplots, as identified in the research gaps in \cref{sec:requirement_analysis}. ParetoTracker enhances these functionalities by providing an integrated visual analytics environment with detailed visualizations of generation sequence and evolutionary operators. This facilitates more effective inspection of individuals and operators across generations, a feature not addressed by the computation toolkits. With the additional design of rich interactions among multiple views, ParetoTracker enables multi-level analysis from quality measures to operators, which assists in identifying notable patterns in evolutionary processes and correlating pattern influences at various granularity.

\vspace{1.2mm}\noindent\textbf{Scalability.} Our current framework effectively manages data logging for standard experimental setups, handling population sizes per generation up to several hundred. Yet, in large-scale optimization scenarios~\cite{Tian2021} with significantly larger dimensionality and population sizes, the log data volume for evolutionary processes can quickly grow. We plan to incorporate advanced sampling strategies prioritizing critical generation subsequences in future implementations. Additionally, the risk of visual clutter in scatterplot-based visualizations requires scatterplot simplification methods~\cite{Yuan2021Evaluation,Sarikaya2018} to address this problem.

\vspace{1.2mm}\noindent\textbf{Generalizability.} The data abstraction framework designed is adaptable to most MOEAs by offering ample detail and ensuring adaptability for various environmental selection processes. However, applying it to decomposition-based MOEAs is challenging due to their complex selection mechanisms that resist straightforward abstraction. For these MOEAs, our current strategy uses a higher-level abstraction that only tracks individuals' survival status. Future work will focus on creating more specialized methods to better support such MOEAs.

\vspace{1.2mm}\noindent\textbf{Limitations and Future Work.} Future enhancements could include a plugin-based environment for easy integration of new evolutionary operators. Additionally, incorporating application-specific information could improve utility, particularly in multiple-criteria decision-making applications without reference points, enabling relative comparison between generations and different algorithm runs. This would allow analysis without ground truth, enhancing the framework's applicability and real-world value. To extend to a broader audience, including educational settings, we plan to assess ParetoTracker's effectiveness across diverse user groups with varying expertise levels.